\documentclass[runningheads]{llncs}
\usepackage[T1]{fontenc}
\usepackage{graphicx}
\usepackage{booktabs}
\usepackage{algorithm}
\usepackage{algpseudocode}
\usepackage{amsmath}
\usepackage{xcolor}
\usepackage[misc]{ifsym}

\usepackage{booktabs}
\usepackage{mwe}

\begin{document}

\title{Divide, Specialize, and Route: A New Approach to Efficient Ensemble Learning}

\titlerunning{Divide, Specialize, and Route}

\author{Jakub Piwko, Jędrzej Ruciński, Dawid Płudowski, Antoni Zajko, Patryzja Żak, Mateusz Zacharecki, Anna Kozak, Katarzyna Woźnica}

\authorrunning{Piwko, Ruciński, Płudowski, Zajko, Żak, Zacharecki}


\institute{Warsaw University of Technology \\
\email{jakub.piwko2.stud@pw.edu.pl, jedrzej.rucinski.stud@pw.edu.pl, dawid.pludowski.stud@pw.edu.pl}}

\maketitle              

\begin{abstract}
Ensemble learning has proven effective in boosting predictive performance, but traditional methods such as bagging, boosting, and dynamic ensemble selection (DES) suffer from high computational cost and limited adaptability to heterogeneous data distributions. To address these limitations, we propose Hellsemble, a novel and interpretable ensemble framework for binary classification that leverages dataset complexity during both training and inference. Hellsemble incrementally partitions the dataset into “circles of difficulty” by iteratively passing misclassified instances from simpler models to subsequent ones, forming a committee of specialised base learners. Each model is trained on increasingly challenging subsets, while a separate router model learns to assign new instances to the most suitable base model based on inferred difficulty. Hellsemble achieves strong classification accuracy while maintaining computational efficiency and interpretability. Experimental results on OpenML-CC18 and Tabzilla benchmarks demonstrate that Hellsemble often outperforms classical ensemble methods. Our findings suggest that embracing instance-level difficulty offers a promising direction for constructing efficient and robust ensemble systems.

\end{abstract}

\section{Introduction}

Lately, classical machine learning has increasingly leaned toward ensemble methods -- techniques that combine multiple learning algorithms to achieve improved predictive performance. These methods focus on training several models on the same dataset and then aggregating their outputs in various ways to make final predictions. Popular examples include bagging, e.g., Random Forests~\cite{breiman2001random}, boosting, e.g., as in XGBoost~\cite{Chen_2016}, voting ensembles~\cite{dietterich2000ensemble}, and strategies used in more complex AutoML systems, e.g., AutoGluon~\cite{erickson2020autogluon}, AutoSklearn~\cite{feurer2022auto}, or TPOT~\cite{olson2016evaluationtreebasedpipelineoptimization}. Ensemble methods often help reduce overfitting, especially when the ensemble consists of diverse models that capture different patterns in the data and can compensate for the weaknesses of other models in the committee.

However, ensemble methods have general drawbacks. They are typically computationally expensive, as they require training multiple models, often on the entire dataset. Moreover, during inference, all component models are usually needed to make a single prediction, resulting in high computational costs.

Another limitation of traditional global ensembles is that they treat the dataset uniformly. Real-world datasets are often complex, exhibiting heterogeneous structures, intricate dependencies, and imbalanced distributions across the feature space. Applying a single model or even a global ensemble can lead to noticing dominant patterns, while subtle yet important relationships may be neglected. In such cases, models may generalise poorly to underrepresented or more difficult regions of the dataset.

To address this issue, \textit{Dynamic Ensemble Selection} (DES) methods have been proposed. These methods train a pool of models on the dataset and, at inference time, dynamically select the most competent model(s) to predict each new instance. Selection is often based on the performance of each model on the local neighbourhood of the query instance, typically identified using nearest neighbours in the feature space.

While DES methods better handle heterogeneous data by leveraging models that specialise in different regions of the feature space, they still incur significant inference-time overhead. This includes finding the neighbourhood of the new instance and maintaining detailed performance records of all models. Additionally, their performance heavily depends on the choice of metric in which the neighbourhood is defined and the model combination strategy. Moreover, all models in DES are still trained on the entire dataset, which may limit their specialisation and miss the opportunity to leverage the structural segmentation of the data.

In this paper, we propose \textbf{Hellsemble} -- a novel ensemble learning framework for binary classification tasks that explicitly utilises dataset complexity in both training and inference while maintaining low computational cost. Unlike traditional ensemble approaches that fit multiple models on the full dataset, Hellsemble incrementally partitions the dataset into subspaces of increasing difficulty during the training process. This is achieved by examining which instances are correctly or incorrectly classified by each model in sequence. Instances misclassified by one simple model are passed to the next simple model in the chain, which attempts to learn their structure more effectively.

This process results in a committee of specialised models, each focusing on a distinct subset of the data. This not only reduces the training overhead but also yields valuable meta-information about the distribution of difficulty across instances in the dataset.

The main contributions of this paper are:
\begin{enumerate}
    \item We introduce \textbf{Hellsemble}, a novel, simple and explainable framework for creating ensemble models for binary classification, combining dataset partitioning and dynamic model selection techniques.
    \item We demonstrate that Hellsemble achieves competitive performance with classical machine learning models on benchmark datasets from OpenML-CC18 and Tabzilla, often outperforming them in terms of classification accuracy.
\end{enumerate}

\section{Related Works}

Ensemble learning has become a widely adopted technique in machine learning due to its ability to improve predictive accuracy by combining multiple models. Classical strategies include bagging, which builds an ensemble by training models independently on bootstrapped samples \cite{cutler2012random}, and boosting \cite{freund1997decision}, which sequentially focuses on correcting the errors of prior models. Another popular approach is voting, where multiple base models cast predictions and the final output is determined by majority or weighted consensus. Stacking takes this a step further by training a meta-model to integrate the predictions of base learners \cite{sesmero2015generating}. More recent AutoML frameworks, such as those found in automated machine learning systems, employ greedy search or optimisation strategies to select and combine models efficiently \cite{caruana2006getting,caruana2004ensemble}. Systems like AutoSklearn~\cite{feurer2022auto} and AutoGluon~\cite{erickson2020autogluontabularrobustaccurateautoml} automate not only model selection but also ensemble construction, often yielding robust and competitive performance with minimal human intervention.

Dynamic model selection aims to improve performance by tailoring the prediction process to each specific instance. Instead of applying a fixed ensemble to all inputs, this approach dynamically selects one or more models based on their estimated competence in the local region of the input space \cite{ko2008dynamic,cavalin2013dynamic}. Typically, this involves analysing the neighbourhood of the query instance, using distance-based methods like k-nearest neighbours, to evaluate how well each model has performed in similar past cases \cite{giacinto2001dynamic,woods1997combination}. The most suitable models are then chosen for that instance, potentially enhancing accuracy in heterogeneous datasets where different regions of the feature space require different inductive biases. Some techniques include special estimation of the competence of each model, which helps in selecting the correct set of predictors~\cite{cruz2015meta}. This adaptiveness allows the system to handle complex data distributions more effectively than static ensembles.

Another promising direction for improving ensemble performance involves routing mechanisms, where input instances are directed to specific models or subsets of models based on learned rules or classifier outputs. These router-based systems often operate by training a separate routing function that learns to assign data points to the models best suited to handle them~\cite{bekku2024stable}. This approach is also used in deep learning, in image classification to create paths to models appropriate to the task~\cite{mcgill2017deciding} and LLMs to move queries to appropriate models based on their difficulty~\cite{ding2024hybrid}. By learning to partition the input space during training, these methods can reduce redundancy in the ensemble and avoid overfitting, while also offering opportunities for parallel inference and scalability.

\section{Hellsemble}

The Hellsemble framework is built upon three key concepts: dynamic model selection, iterative (sequential from a predefined set or greedy) model addition, and router-based prediction routing. Specifically, Hellsemble selects only a single model from its committee to make a prediction for each new instance, making it a dynamic selection method. It uses a greedy strategy during training by selecting, at each iteration, the model that provides the greatest improvement in validation performance. Additionally, a dedicated router model is trained to determine which base model from the committee should handle a given input during inference. To mitigate overfitting, Hellsemble evaluates model performance on a validation subset throughout training and applies a regularization strategy based on overfitting control. The framework is implemented in Python, and users must provide a list of candidate base models and a router model.

\begin{figure}[h!]
  \centering
  \includegraphics[width=\textwidth]{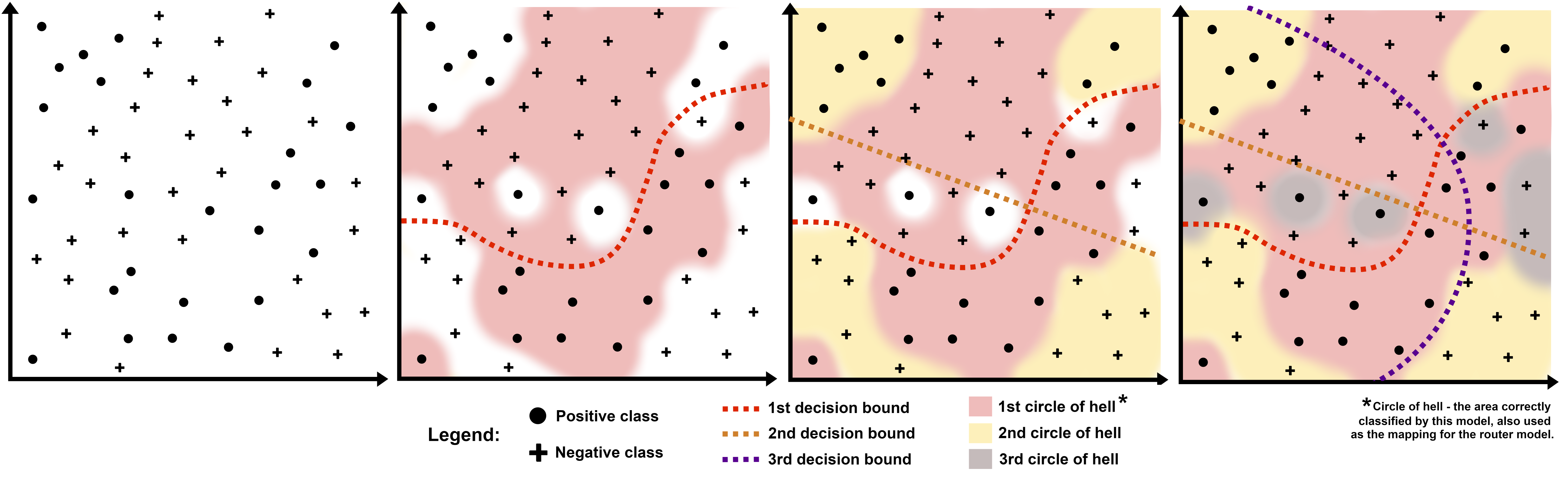}
  \caption{Training and dataset evalution schema for Hellsemble}
  \label{fig:train_schema}
\end{figure}

The training procedure and evolution of the dataset across iterations are illustrated in Figure~\ref{fig:train_schema}, which we refer to throughout this section.

Hellsemble builds what we refer to as ``circles of difficulty'' - based on how well models perform on the data in each iteration. In every round, the algorithm selects the model which, when added to the current Hellsemble committee, results in the highest improvement on the validation score. This score can be any evaluation metric, such as accuracy or AUC, or even a user-defined function.

Once the best model is selected in a given iteration, we analyse which observations it predicted correctly and incorrectly. In the basic version, only the incorrectly classified instances are passed to the next iteration. These are the more challenging examples, and the goal is to find another model that can handle them better. In Figure~\ref{fig:train_schema}, this is visualised as overlapping colored regions corresponding to decision boundaries of different models.

To avoid overfitting and allow overlap between difficulty levels, we also include a portion of correctly classified examples in the next iteration. The number of such examples is determined based on the performance difference between the training and validation datasets—specifically, to minimize the gap between them. This acts as a form of regularisation, ensuring that each model generalises beyond a narrow region.

After each iteration, the router model is retrained. Based on the accumulated difficulty information -- the iteration at which each instance was filtered out -- we assign instances to classes representing their ``circles of difficulty''. This effectively transforms the problem into a multi-class classification task where the router learns to predict which base model should be used for a given input. The router model thus captures the segmentation of the dataset into regions of varying difficulty.

This iterative process continues until either there are no incorrectly classified instances left or no additional model improves the validation score. At the end of training, we obtain a list of base models, each trained on increasingly difficult subsets of the data, and a router model that maps new instances to the appropriate base model.

The key advantage of Hellsemble lies in its simplicity and efficiency. Since each base model is trained only on a specific subset of the data, they do not need to be highly general models -- simple models are sufficient if they are accurate within their targeted region. Conversely, the router model must be expressive enough to capture nonlinear boundaries, as it performs a multiclass classification task that determines which model to delegate prediction to. This structure makes Hellsemble an interpretable and efficient ensemble learning strategy.

\subsection{Sequential Hellsemble}

Sequential Hellsemble treats the order of user-specified models as the fixed order in which models are trained and added to successive circles of difficulty. Its details are pointed out in Algorithm~\ref{alg:seq}.

\begin{algorithm}[H]
\caption{Sequential Hellsemble}
\label{alg:seq}
\begin{algorithmic}[1]
\Require Dataset $D$, base models $M = \{M_1, M_2, \ldots, M_k\}$, router model $R$
\State Split $D$ into $D_{\text{train}}$ and $D_{\text{val}}$
\State $E \gets \emptyset$ \Comment{Empty ensemble}
\State $D_{\text{router}} \gets \emptyset$
\State $S_{\text{prev}} \gets 0$
\For{$i = 1$ to $k$}
    \State Train $M_i$ on $D_{\text{train}}$
    \State $E \gets E \cup \{M_i\}$
    \State Label each instance in $D_{\text{train}}$ with class $i$
    \State $D_{\text{router}} \gets D_{\text{router}} \cup D_{\text{train}}$
    \State $R \gets$ train router on $D_{\text{router}}$ (as multiclass classifier)
    \State $S_{\text{curr}} \gets$ evaluate $E$ on $D_{\text{val}}$ using $R$
    \If{$S_{\text{curr}} \leq S_{\text{prev}}$}
        \State \textbf{break}
    \Else
        \State $S_{\text{prev}} \gets S_{\text{curr}}$
    \EndIf
    \State $D_{\text{hard}} \gets$ instances in $D_{\text{train}}$ misclassified by $M_i$
    \State $D_{\text{easy}}^{(\alpha)} \gets$ a fraction $\alpha$ of correctly predicted instances
    \State $D_{\text{train}} \gets D_{\text{hard}} \cup D_{\text{easy}}^{(\alpha)}$
\EndFor
\State \Return $E$, $R$
\end{algorithmic}
\end{algorithm}

\subsection{Greedy Hellsemble}

The Greedy Hellsemble is a more computationally intensive variant of the Sequential Hellsemble. Rather than using a fixed order of models, it dynamically selects the most promising model in each iteration by evaluating all available candidates.

The core algorithm remains identical to the Sequential Hellsemble, but with the following key changes:

\begin{itemize}
    \item \textbf{Model Evaluation at Each Iteration:} Instead of selecting the next model in a predefined order, all models in the list are trained on the current training set in each iteration.
    
    \item \textbf{Greedy Selection Based on Validation Performance:} Each trained model is temporarily added to the ensemble and evaluated on the validation set. Only the model that results in the best performance improvement is selected and permanently added to the ensemble.
\end{itemize}

The remaining steps -- including router training, training set updates, regularization, and stopping criteria -- are unchanged from Algorithm~\ref{alg:seq}.

\vspace{0.5em}
\noindent
The following pseudocode replaces lines 5-7 of Algorithm~1 in the greedy variant:

\begin{algorithm}[H]
\caption{Greedy Model Selection (replaces lines 5--7 of Algorithm~\ref{alg:seq})}
\begin{algorithmic}[1]
\State $best\_score \gets -\infty$
\State $best\_model \gets \text{None}$
\For{each model $M_j$ in model list}
    \State $M_j \gets$ train on $D_{\text{train}}$
    \State $E \gets E \cup \{M_j\}$ \Comment{Temporarily add to ensemble}
    \State $score_j \gets$ evaluate $E$ on $D_{\text{val}}$
    \If{$score_j > best\_score$}
        \State $best\_score \gets score_j$
        \State $best\_model \gets M_j$
    \EndIf
    \State $E \gets E \setminus \{M_j\}$ \Comment{Remove temporary model}
\EndFor
\State $E \gets E \cup \{best\_model\}$ \Comment{Permanently add best model}
\end{algorithmic}
\end{algorithm}

\section{Experiments}

To evaluate the performance of the proposed Hellsemble framework, we conducted extensive experiments using two collections of benchmark binary classification datasets: \textbf{OpenML-CC18}~\cite{bischl2021openmlbenchmarkingsuites} and \textbf{Tabzilla}~\cite{mcelfresh2024neuralnetsoutperformboosted}. These datasets vary in complexity, with Tabzilla offering more challenging structures and higher feature dimensionality. Together, they provide a robust suites for assessing model generalisation and adaptability.

The experimental setup was designed to investigate the effect of different combinations of base models and router models on classification performance. We selected:
\begin{itemize}
    \item 4 distinct configurations of base models (i.e., different sets of classifiers),
    \item 4 different router models.
\end{itemize}
Each base model configuration was paired with each router model exactly once, resulting in a total of \textbf{16 unique Hellsemble configurations}. The Table~\ref{tab:experiment} summarizes the combinations of base models and router models used in our experiments.

\begin{table}[H]
\centering
\caption{Summary of base and router model combinations used in experiments.}
\label{tab:experiment}
\begin{tabular}{p{9cm} p{4cm}}
\toprule
\textbf{Base Models} & \textbf{Router Model} \\
\midrule
KNN, Logistic Regression, Decision Tree, GaussianNB & KNN ($k$=3) \\
 & KNN ($k$=5) \\
 & MLP \\
 & Random Forest \\
\midrule
Random Forest, XGBoost, MLP & KNN ($k$=3) \\
 & KNN ($k$=5) \\
 & MLP \\
 & Random Forest \\
\midrule
KNN ($k$=3), KNN ($k$=5), Decision Tree, GaussianNB & KNN ($k$=3) \\
 & KNN ($k$=5) \\
 & MLP \\
 & Random Forest \\
\midrule
All models (KNNs, Decision Tree, GaussianNB, RFs, XGBoost, MLPs) & KNN ($k$=3) \\
 & KNN ($k$=5) \\
 & MLP \\
 & Random Forest \\
\bottomrule
\end{tabular}
\end{table}


For each dataset, we performed the following steps:
\begin{enumerate}
    \item The dataset was split into a training and a test subset using stratified sampling.
    \item Each of the 16 Hellsemble configurations was trained using both the \textbf{sequential} and the \textbf{greedy} variant of the algorithm. Note that due to the nature of the Hellsemble framework, the number of models included in each ensemble may vary depending on the iteration process.
    \item To serve as a baseline, all individual base models used in the ensemble configurations were also trained separately on the training set.
    \item Finally, both the trained Hellsemble models and individual base models were evaluated on the test set. This allowed for a direct comparison of ensemble performance versus standalone classifiers.
\end{enumerate}

The goal of this experimental evaluation is to assess whether Hellsemble -- under varying configurations -- offers consistent improvements over individual models, and to observe how its performance differs between the sequential and greedy variants.

\section{Results}

This section presents a summary of the experimental results. Our primary objective is to compare the performance of the base models with the proposed Hellsemble models — both the sequential and greedy variants — across various configurations. For each combination of base model suite and router, we assess performance using the average accuracy metric.

\begin{figure}[h!]
  \centering
  \includegraphics[width=\textwidth]{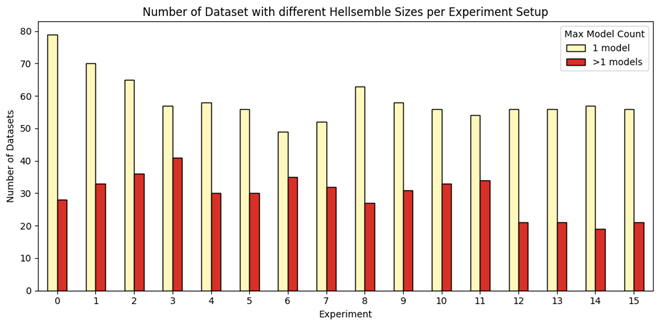}
  \caption{Proportion of experiments in each configuration where the Hellsemble (either sequential or greedy) consisted of exactly one model versus more than one model.}
  \label{fig:dataset_ratio}
\end{figure}

Before analyzing performance metrics, we first examine how frequently Hellsemble constructs ensembles consisting of more than one model. Importantly, we did not enforce multi-model construction during training—Hellsemble naturally stops adding models when the inclusion of an additional model fails to improve validation performance. This means single-model Hellsembles are valid and expected in certain scenarios.

Figure~\ref{fig:dataset_ratio} shows that in most configurations, Hellsemble tends to consist of only a single model. Particularly, configurations using the most extensive base model suites (the final four setups) resulted in the lowest proportion of multi-model Hellsembles. On the other hand, configurations 3 and 4 produced the most diverse Hellsembles in terms of model count.

For the following accuracy analysis, we restrict our attention to those dataset and experiment pairs where the resulting Hellsemble (either sequential or greedy) included more than one model. This focus allows us to evaluate whether the multi-model nature of Hellsemble contributes to performance improvements over individual base models.

To that end, we aggregated accuracy scores for all models (base models and Hellsemble variants) across all selected datasets and configurations. The results are visualized in the following plots, grouped by base model suite and routing model.

\begin{figure}[h!]
  \centering
  \includegraphics[width=\textwidth]{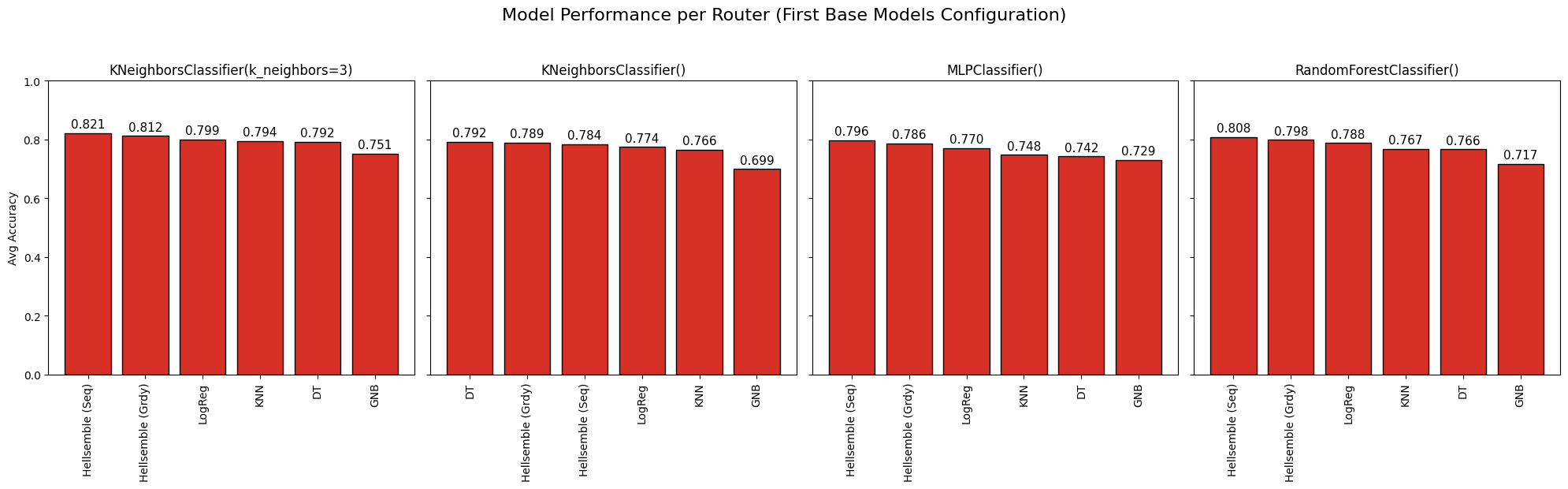}
  \caption{Average accuracy across all datasets for different routing models and the first base model suite (e.g., KNN, Naive Bayes, Logistic Regression, and Decision Trees).}
  \label{fig:avg_acc_1}
\end{figure}

Figure~\ref{fig:avg_acc_1} presents results for configurations using classic, simple models such as KNN, Gaussian Naive Bayes, Logistic Regression, and Decision Trees. In nearly all routing scenarios, Hellsemble outperforms the individual base models in terms of average accuracy. The one exception is when using KNN with default settings as the router—here, the Decision Tree performs best. Nevertheless, these results indicate that even with basic classifiers, Hellsemble can yield accuracy improvements.

\begin{figure}[h!]
  \centering
  \includegraphics[width=\textwidth]{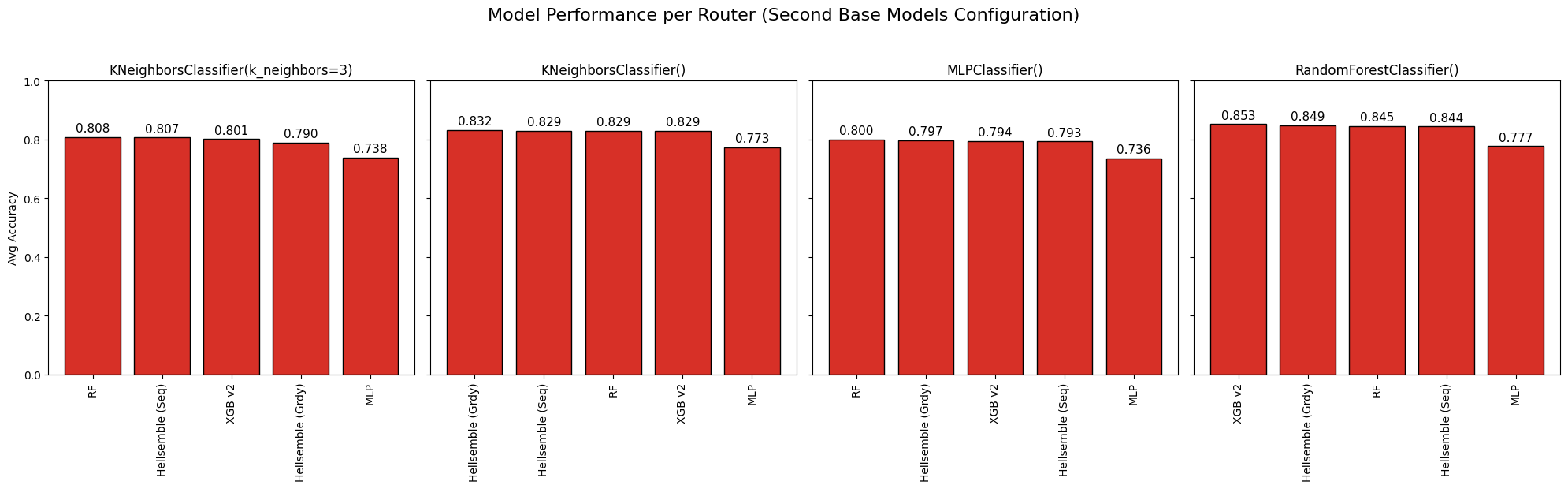}
  \caption{Average accuracy across all datasets for different routing models and the second base model suite (e.g., Random Forests and XGBoost).}
  \label{fig:avg_acc_2}
\end{figure}

Figure~\ref{fig:avg_acc_2} compares performance in configurations with more advanced models: Random Forests and XGBoost, both of which are strong ensemble learners for tabular data. In these setups, Hellsemble performs slightly less favorably. In several configurations, such as those using Decision Trees or MLPs as routers, XGBoost outperforms all other models. Only in one configuration (the second router) does Hellsemble (greedy) achieve the best result.

These findings suggest that when the base models are already powerful ensemble methods, Hellsemble may struggle to deliver further gains. This could also reflect the router’s limited effectiveness in correctly assigning instances to appropriate difficulty "circles" when all base models are inherently strong. However, Hellsemble is primarily designed for scenarios where simpler models can specialize on subsets of the data—an advantage that becomes less relevant when base models already generalize well.

\begin{figure}[h!]
  \centering
  \includegraphics[width=\textwidth]{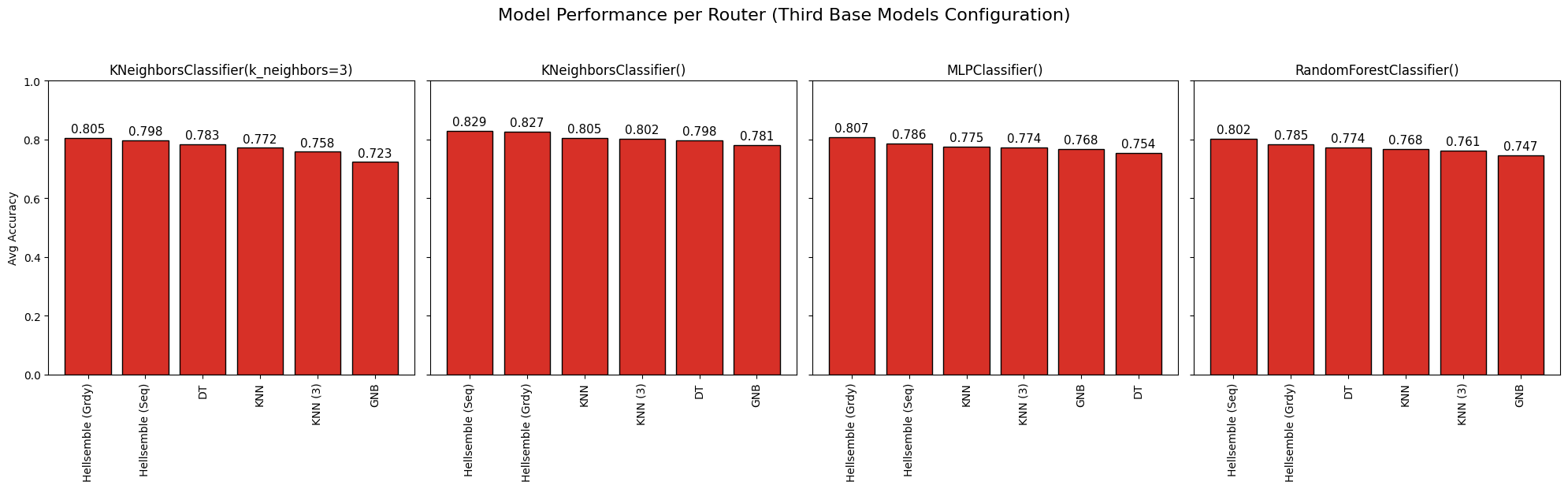}
  \caption{Average accuracy across all datasets for different routing models and the third base model suite (composed of simpler classifiers).}
  \label{fig:avg_acc_3}
\end{figure}

In Figure~\ref{fig:avg_acc_3}, we analyze results from a suite consisting entirely of simpler models. In this case, Hellsemble consistently outperforms all base models across all router types. These results support the hypothesis that Hellsemble is particularly well-suited to settings involving lightweight classifiers. The observed performance improvements of 2–3 percentage points are achieved without adding substantial complexity —only by routing samples to more specialized models.

\begin{figure}[h!]
  \centering
  \includegraphics[width=\textwidth]{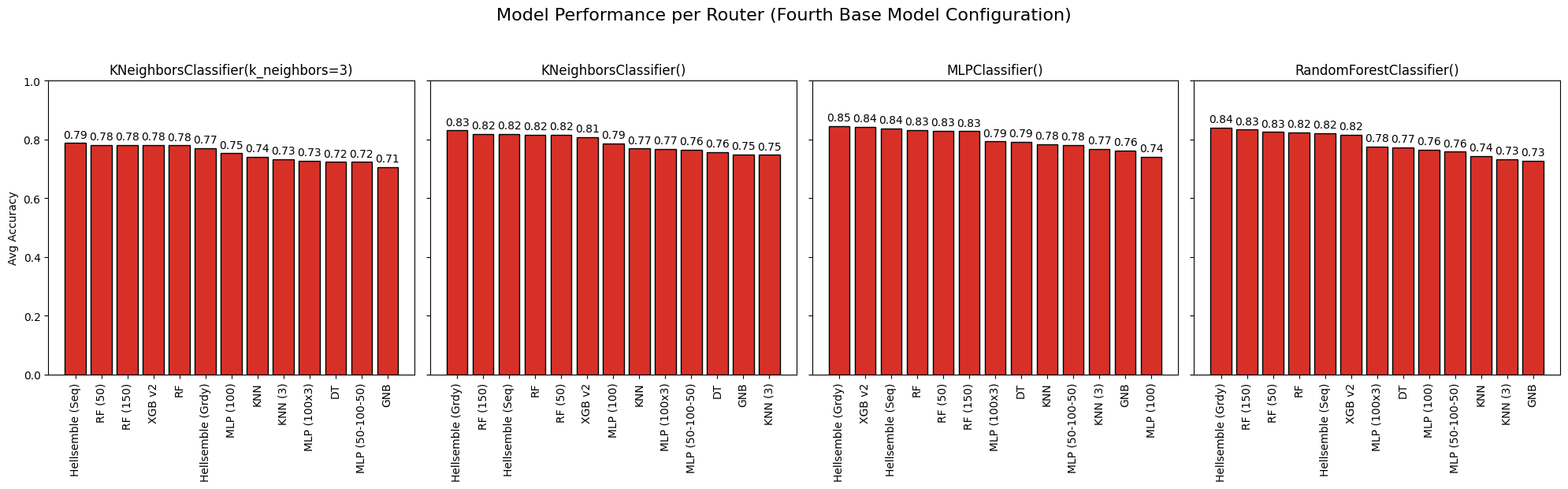}
  \caption{Average accuracy across all datasets for different routing models and the fourth base model suite (containing both simple and advanced models).}
  \label{fig:avg_acc_4}
\end{figure}

Finally, Figure~\ref{fig:avg_acc_4} shows results for the most extensive base model suite, comprising 11 classifiers, including basic models, Decision Trees, XGBoost, and MLPs. Despite the strength of the individual models, Hellsemble still manages to outperform all base models in every router configuration. Sequential Hellsemble performs best with KNN (3 neighbors) as the router, while greedy Hellsemble dominates in the other setups.

This result is particularly promising: even when the base pool includes powerful models like XGBoost and Random Forests, Hellsemble is able to form specialized ensembles that achieve superior performance. Although the improvements are modest, they validate the premise of Hellsemble—that leveraging model specialization based on data subregions can enhance overall accuracy.

It is worth noting that in this configuration, fewer datasets resulted in multi-model Hellsembles. However, when such ensembles are formed, their performance demonstrates the potential of the Hellsemble architecture. These findings suggest that model specialization through difficulty-aware routing is a viable strategy for boosting classifier performance, especially when using simpler models.

\section{Conclusion}

Hellsemble is a novel ensemble learning framework that combines concepts from dynamic ensemble selection, greedy model construction, and routing-based instance selection. It introduces a simple yet effective strategy for building specialized ensembles by sequentially training base models on progressively harder subsets of the data—referred to as circles of difficulty—defined by the misclassified instances of previous models.

This design makes Hellsemble both interpretable and efficient. By utilizing lightweight models and training them on smaller, focused subsets, the framework achieves competitive performance while keeping computational costs low. At inference time, only the router and a single specialized model are involved in prediction, reducing latency compared to traditional ensemble methods.

Experimental results on two benchmark dataset collections and a range of model configurations demonstrate that Hellsemble can outperform its base models in many settings, particularly when the base learners are simple classifiers. Even in cases where advanced models like XGBoost and Random Forests are included in the base suite, Hellsemble is often able to match or exceed their performance.

These results highlight the potential of Hellsemble as a practical and scalable ensemble method. While the current implementation already achieves promising results, there remains room for further improvement—particularly in enhancing the router’s accuracy and exploring alternative strategies for defining and managing circles of difficulty.

In summary, Hellsemble offers a compelling new direction in ensemble learning, balancing simplicity, efficiency, and predictive performance through a modular and interpretable architecture.

\begin{credits}
\subsubsection{\ackname} Work on this project is financially supported by the Warsaw University of Technology as a part of the \textit{Initiative: Research University (IDUB) programme.} program, through a grant program for scientific clubs for the years 2024–2025.

\end{credits}

\begin{thebibliography}{10}
\providecommand{\url}[1]{\texttt{#1}}
\providecommand{\urlprefix}{URL }
\providecommand{\doi}[1]{https://doi.org/#1}

\bibitem{bekku2024stable}
Bekku, H., Kume, T., Tsuge, A., Nakazawa, J.: A stable and efficient dynamic ensemble method for pothole detection. Pervasive and Mobile Computing  \textbf{104},  101973 (2024)

\bibitem{bischl2021openmlbenchmarkingsuites}
Bischl, B., Casalicchio, G., Feurer, M., Gijsbers, P., Hutter, F., Lang, M., Mantovani, R.G., van Rijn, J.N., Vanschoren, J.: Openml benchmarking suites (2021), \url{https://arxiv.org/abs/1708.03731}

\bibitem{breiman2001random}
Breiman, L.: Random forests. Machine learning  \textbf{45},  5--32 (2001)

\bibitem{caruana2006getting}
Caruana, R., Munson, A., Niculescu-Mizil, A.: Getting the most out of ensemble selection. In: Sixth International Conference on Data Mining (ICDM'06). pp. 828--833. IEEE (2006)

\bibitem{caruana2004ensemble}
Caruana, R., Niculescu-Mizil, A., Crew, G., Ksikes, A.: Ensemble selection from libraries of models. In: Proceedings of the twenty-first international conference on Machine learning. p.~18 (2004)

\bibitem{cavalin2013dynamic}
Cavalin, P.R., Sabourin, R., Suen, C.Y.: Dynamic selection approaches for multiple classifier systems. Neural computing and applications  \textbf{22},  673--688 (2013)

\bibitem{Chen_2016}
Chen, T., Guestrin, C.: Xgboost: A scalable tree boosting system. In: Proceedings of the 22nd ACM SIGKDD International Conference on Knowledge Discovery and Data Mining. p. 785–794. KDD ’16, ACM (Aug 2016). \doi{10.1145/2939672.2939785}, \url{http://dx.doi.org/10.1145/2939672.2939785}

\bibitem{cruz2015meta}
Cruz, R.M., Sabourin, R., Cavalcanti, G.D., Ren, T.I.: Meta-des: A dynamic ensemble selection framework using meta-learning. Pattern recognition  \textbf{48}(5),  1925--1935 (2015)

\bibitem{cutler2012random}
Cutler, A., Cutler, D.R., Stevens, J.R.: Random forests. Ensemble machine learning: Methods and applications pp. 157--175 (2012)

\bibitem{dietterich2000ensemble}
Dietterich, T.G.: Ensemble methods in machine learning. In: International workshop on multiple classifier systems. pp. 1--15. Springer (2000)

\bibitem{ding2024hybrid}
Ding, D., Mallick, A., Wang, C., Sim, R., Mukherjee, S., Ruhle, V., Lakshmanan, L.V., Awadallah, A.H.: Hybrid llm: Cost-efficient and quality-aware query routing. arXiv preprint arXiv:2404.14618  (2024)

\bibitem{erickson2020autogluon}
Erickson, N., Mueller, J., Shirkov, A., Zhang, H., Larroy, P., Li, M., Smola, A.: Autogluon-tabular: Robust and accurate automl for structured data (2020), \url{https://arxiv.org/abs/2003.06505}

\bibitem{erickson2020autogluontabularrobustaccurateautoml}
Erickson, N., Mueller, J., Shirkov, A., Zhang, H., Larroy, P., Li, M., Smola, A.: Autogluon-tabular: Robust and accurate automl for structured data (2020), \url{https://arxiv.org/abs/2003.06505}

\bibitem{feurer2022auto}
Feurer, M., Eggensperger, K., Falkner, S., Lindauer, M., Hutter, F.: Auto-sklearn 2.0: Hands-free automl via meta-learning. Journal of Machine Learning Research  \textbf{23}(261),  1--61 (2022)

\bibitem{freund1997decision}
Freund, Y., Schapire, R.E.: A decision-theoretic generalization of on-line learning and an application to boosting. Journal of computer and system sciences  \textbf{55}(1),  119--139 (1997)

\bibitem{giacinto2001dynamic}
Giacinto, G., Roli, F.: Dynamic classifier selection based on multiple classifier behaviour. Pattern Recognition  \textbf{34}(9),  1879--1881 (2001)

\bibitem{ko2008dynamic}
Ko, A.H., Sabourin, R., Britto~Jr, A.S.: From dynamic classifier selection to dynamic ensemble selection. Pattern recognition  \textbf{41}(5),  1718--1731 (2008)

\bibitem{mcelfresh2024neuralnetsoutperformboosted}
McElfresh, D., Khandagale, S., Valverde, J., C, V.P., Feuer, B., Hegde, C., Ramakrishnan, G., Goldblum, M., White, C.: When do neural nets outperform boosted trees on tabular data? (2024), \url{https://arxiv.org/abs/2305.02997}

\bibitem{mcgill2017deciding}
McGill, M., Perona, P.: Deciding how to decide: Dynamic routing in artificial neural networks. In: International Conference on Machine Learning. pp. 2363--2372. PMLR (2017)

\bibitem{olson2016evaluationtreebasedpipelineoptimization}
Olson, R.S., Bartley, N., Urbanowicz, R.J., Moore, J.H.: Evaluation of a tree-based pipeline optimization tool for automating data science (2016), \url{https://arxiv.org/abs/1603.06212}

\bibitem{sesmero2015generating}
Sesmero, M.P., Ledezma, A.I., Sanchis, A.: Generating ensembles of heterogeneous classifiers using stacked generalization. Wiley interdisciplinary reviews: data mining and knowledge discovery  \textbf{5}(1),  21--34 (2015)

\bibitem{woods1997combination}
Woods, K., Kegelmeyer, W.P., Bowyer, K.: Combination of multiple classifiers using local accuracy estimates. IEEE transactions on pattern analysis and machine intelligence  \textbf{19}(4),  405--410 (1997)

\end{thebibliography}
\end{document}